\documentclass{article}

\usepackage[table]{xcolor}
\usepackage[preprint]{corl_2026}
\usepackage{amsmath,amssymb, amsthm, graphicx, caption, subcaption, wrapfig, algorithm, algpseudocode, booktabs, multirow, mathtools}
\usepackage[belowfloats]{footmisc} 
\theoremstyle{definition}
\newtheorem{definition}{Definition}

\title{MAPO\@: Mixture of Experts with Alternating Policy Optimization for Universal Humanoid Motion Tracking}

\author{
  \begin{NoHyper}
  Seonghyeon Lim\textsuperscript{1} \quad
  Hyeonwoo Lee\textsuperscript{1} \quad
  Mincheol Kim\textsuperscript{2} \quad
  Kimin Lee\textsuperscript{3,4}\thanks{Corresponding authors.} \quad
  Hyun Myung\textsuperscript{1}\footnotemark[2]
  \end{NoHyper} \\
  \textsuperscript{1}School of Electrical Engineering, KAIST \quad
  \textsuperscript{2}Robotics Program, KAIST \\
  \textsuperscript{3}Graduate School of AI, KAIST \quad
  \textsuperscript{4}Config \\
  \texttt{\{shlim, hyeonwoolee, kmc96, kiminlee, hmyung\}@kaist.ac.kr}
}

\begin{document}
\makeatletter
\renewcommand{\@notice}{}%
\makeatother
\setcounter{footnote}{1} 
\maketitle
\setcounter{footnote}{0}
\vspace{-10mm}
\begin{abstract}
Universal humanoid motion tracking is essential for enabling humanoid robots to perform diverse tasks in human environments.
To capture the wide diversity of human motions, mixture-of-experts (MoE) policies have been widely adopted, leveraging dynamic routing to enhance policy expressiveness.
However, we identify that divergence between old and updated policies can be misestimated during policy-gradient updates of MoE policies, due to their dynamic routing.
Because most policy-gradient algorithms regulate policy updates based on the divergence, this misestimation can lead to either overly conservative or excessively large updates, which can yield a suboptimal policy.
In this paper, we propose Mixture of Experts with Alternating Policy Optimization (MAPO), which alternately optimizes the router and the experts to avoid the divergence misestimation.
MAPO consistently tracks 182 motions that PPO with MoE fails to track, whereas PPO with MoE tracks only one motion that MAPO fails to track.
MAPO also attains the highest tracking success rate among the compared methods and the lowest tracking error across most error metrics, demonstrating its superior performance.
Videos can be found on the project website: \url{https://mapo-corl2026.github.io/MAPO/}
\end{abstract}
\vspace{-5mm}
\begin{figure}[b!]
    \centering
    \captionsetup{font=small}
    \vspace{-5mm}
    \includegraphics[width=0.95\linewidth]{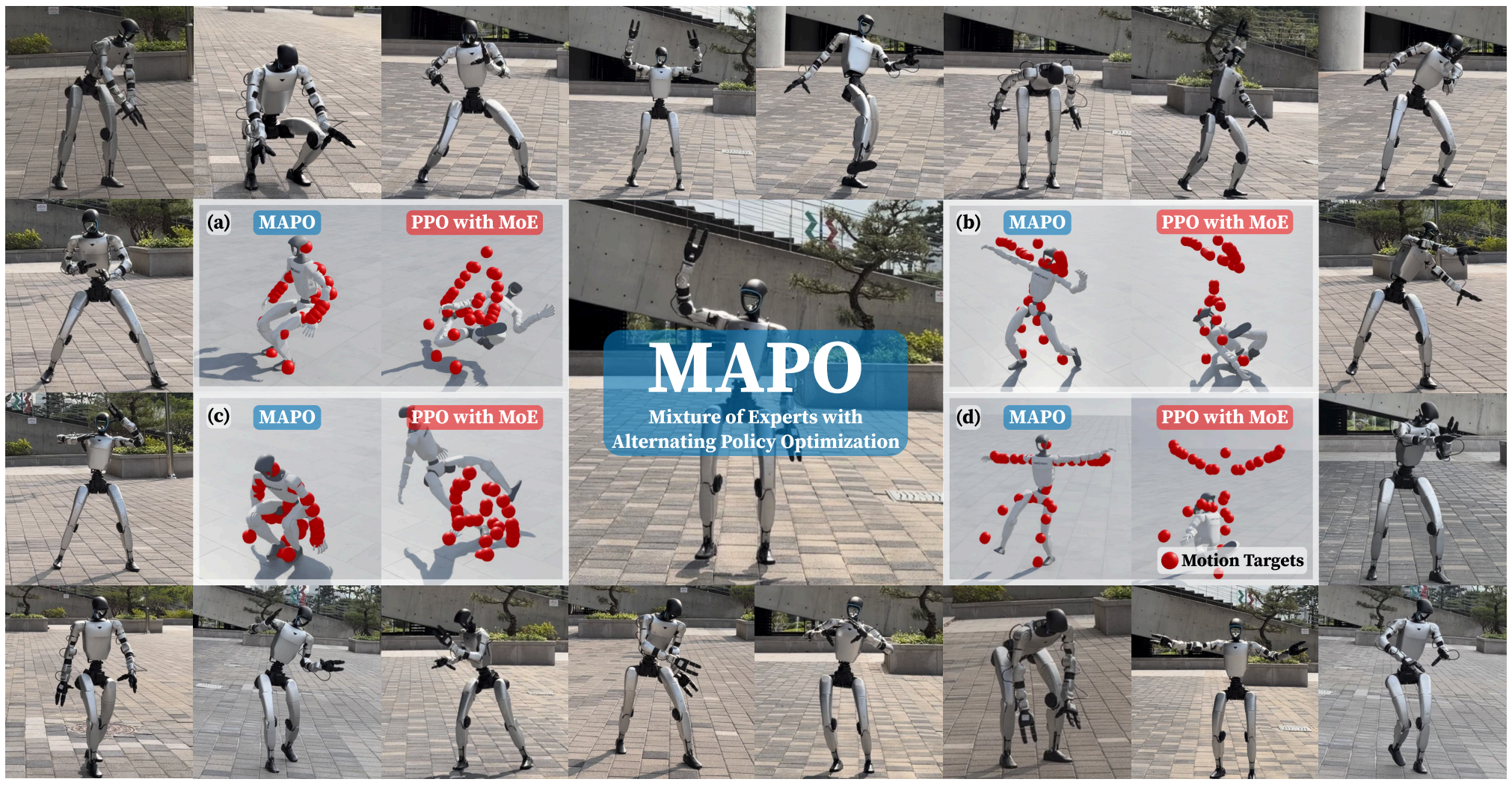}
    \caption{
MAPO resolves the divergence misestimation problem that hinders policy-gradient training of MoE policies, enabling effective humanoid motion tracking.
Consequently, MAPO successfully tracks a wide variety of human motions, including complex whole-body motions, such as (a)~sitting in mid-air, (b)~abrupt full-body turns, (c)~deep crouching, and (d)~one-footed spinning. 
Red spheres denote the reference motion targets. 
    }
    \label{fig:title}
    \vspace{-5mm}
\end{figure}
\section{Introduction}
\label{sec:introduction}
Humanoid robots have the potential to perform various tasks in human environments, given their human-like morphology and whole-body dexterity for locomotion and manipulation.
To achieve such potential, a humanoid should be able to perform diverse motions ranging from basic locomotion to complex whole-body behaviors.
Reinforcement learning (RL)-based approaches have enabled humanoids to perform robust walking~\citep{radosavovic2024real, long2025learning, gu2024advancing, radosavovic2024learning}, parkour~\citep{zhuang2025humanoid, zhu2026hiking}, and fall recovery~\citep{huang2025learning, he2025learning}.
However, extending such approaches to universal humanoid motion tracking requires physically feasible and diverse reference motions during training.
Large-scale human motion datasets~\citep{mahmood2019amass, harvey2020robust, li2023object} have become promising sources of such reference motions~\citep{peng2018deepmimic, cheng2024expressive, ji2025exbody2, liao2025beyondmimic, ze2025twist, luo2025sonic, he2024learning}.

Nevertheless, learning a universal motion tracker with a monolithic policy can be challenging~\citep{chen2025gmt, wang2026general}.
Because diverse reference motions require distinct control strategies, this task can be cast as multi-task learning, where monolithic policies suffer from inter-task interference~\citep{yu2020gradient, liu2021conflict}. 
Moreover, different motion modes often share overlapping state regions~\citep{chen2025gmt}, likely exacerbating the interference~\citep{schaul2019ray}.
In contrast, the mixture of experts (MoE) architecture~\citep{jacobs1991adaptive} has multiple experts and a router as its components, with the router computing mixing weights over the experts for a given state.
This architecture has shown effectiveness in such multi-task learning problems~\citep{willi2024mixture, hendawy2024multi, cheng2023multi, yang2020multi, huang2025moe}.
The MoE architecture has also been explored for motion tracking tasks~\citep{chen2025gmt, wang2026general, han2025kungfubot2, li2025clone}.
However, these studies primarily focus on architectural design, and the problems that arise when MoE policies are trained with widely used policy-gradient algorithms~\citep{schulman2015trust, schulman2017proximal} remain largely unexplored.

In this work, we identify a phenomenon termed \textit{divergence misestimation} where the divergence between the data-collection and updated policies does not accurately reflect the magnitude of MoE policy updates.
Specifically, the phenomenon becomes problematic for algorithms that constrain policy updates via this divergence, including proximal policy optimization (PPO)~\citep{schulman2017proximal}, which is widely adopted in robot control tasks.
For monolithic policies, the magnitude of the policy update is directly reflected in the divergence.
However, the distribution of an MoE policy is a weighted sum of expert outputs, with mixing weights determined by the router.
Therefore, the divergence can change not only with updates to the experts but also with shifts in the mixing weights.
It can thus either overestimate or underestimate the changes in each component, leading to overly conservative or aggressive policy updates, respectively.

To address the divergence misestimation, we propose Mixture of Experts with Alternating Policy Optimization (MAPO).
A naive approach to resolve the divergence misestimation is to freeze the mixing weights at their data-collection values during the update.
Although this approach can avoid the misestimation, it prevents the router from being optimized via the RL objective.
MAPO resolves this trade-off by symmetrically alternating updates of the two components.
During the expert update, the mixing weights are held fixed at their rollout values, and vice versa for the router update.
In this way, MAPO measures the divergence only for the component under update, rather than over the entire MoE policy as in existing policy-gradient algorithms.
Such separation avoids the misestimation, and the alternating updates allow both components to be optimized with the RL objective.

Our contributions are summarized as follows.
(i) We identify divergence misestimation in MoE policy optimization. 
Because most policy-gradient algorithms regulate policy divergence, the misestimation can lead to overly conservative or aggressive updates.
(ii) We propose MAPO, a simple but efficient remedy for the divergence misestimation that alternates expert and router updates, measuring divergence per component.
In our experiments on humanoid motion tracking, the proposed method achieves substantially higher reward and better tracking performance than a PPO-trained MoE policy by avoiding the divergence misestimation (see Figure~\ref{fig:title}).
	

\section{Related Work}
\label{sec:related_works}
\paragraph{Mixture of Experts in Reinforcement Learning:}
Originally proposed by~\citet{jacobs1991adaptive}, MoE has gained increasing attention for scaling up large language models~\citep{shazeer2017outrageously, lepikhin2021gshard, fedus2022switch, zhou2022mixture} and vision models~\citep{riquelme2021scaling}.
MoE has also been explored in RL~\citep{ren2021probabilistic, akrour2021continuous, peng2016terrain}, particularly in multi-task settings~\citep{willi2024mixture, hendawy2024multi, cheng2023multi, yang2020multi, huang2025moe}.
Many of these works adopt a sparse MoE, which activates only a subset of experts for each input, enabling the network to scale efficiently.
However, sparse routing is non-differentiable, hindering end-to-end optimization~\citep{ren2021probabilistic}.
In contrast, dense MoE~\citep{puigcerver2024sparse, yang2019condconv, tian2020conditional, muqeeth2024soft} uses a weighted combination of all experts with continuous mixing weights, making it fully differentiable~\citep{puigcerver2024sparse}.
Recent empirical studies~\citep{willi2024mixture, obando2024mixtures} have analyzed MoE in RL, reporting that differentiable MoE shows better performance than sparse MoE.
Despite these advances, the divergence misestimation phenomenon has remained unaddressed, which we identify and resolve in this work.

\paragraph{Universal Whole-body Control for Humanoids:}
RL-based approaches have achieved robust whole-body control for quadrupeds~\citep{jeon2023learning, he2024quadlearning} and legged manipulators~\citep{fu2023deep, portela2025whole, lim2026openheart, lim2024learning, vijayan2025multi}.
Such approaches require sampling motion targets during training, which is straightforward when motion targets can be abstracted as low-dimensional states, such as body velocities or end-effector poses.
However, this approach is hard to extend to humanoids, where such sampling rarely yields physically feasible whole-body motions.
Instead, DeepMimic~\citep{peng2018deepmimic} demonstrates that human motion data can serve as feasible and diverse reference motions for various simulated avatars. 
This paradigm has been scaled to achieve universal humanoid whole-body control in the real world~\citep{peng2018deepmimic, cheng2024expressive, ji2025exbody2, liao2025beyondmimic, ze2025twist, luo2025sonic, he2024learning, luo2023perpetual, he2025omnih2o, he2025hover, li2025amo, ze2025twist2, araujo2025retargeting}.
However, the increasing scale and diversity of motion data raise the need for more expressive policy networks.
MoE architectures have been explored to improve expressiveness via expert specialization~\citep{chen2025gmt, wang2026general, han2025kungfubot2, li2025clone}.
CLONE~\citep{li2025clone} and KungfuBot2~\citep{han2025kungfubot2} adopt a sparse MoE, whose non-differentiable routing hinders end-to-end RL optimization.
In contrast, GMT~\citep{chen2025gmt} and FAST~\citep{wang2026general} employ a fully differentiable dense MoE.
However, these works primarily focus on architectural design, leaving the effective optimization of MoE policies under widely used policy-gradient algorithms~\citep{schulman2015trust, schulman2017proximal} underexplored.

\section{Background: Mixture of Experts with Policy Gradient}
\label{sec:preliminary}
\subsection{Reinforcement Learning}
RL considers a sequential decision-making problem within a given environment~\citep{sutton1998introduction}.
The environment is typically modeled as a Markov decision process defined by the tuple $\langle \mathcal{S}, \mathcal{A}, P, R, \gamma \rangle$, where $\mathcal{S}$ is the state space, $\mathcal{A}$ is the action space, $P(s'|s,a)$ is the transition probability from state $s$ to $s'$ given action $a$, $R(s,a)$ is the reward function, and $\gamma \in [0,1]$ is the discount factor.
A stochastic policy $\pi(a|s)$ defines a distribution over actions conditioned on the current state. 
Because the desired behavior is encoded in the reward function, the goal of RL is to find an optimal policy $\pi^*$ that maximizes the expected return, $\mathbb{E}_{\pi} \left[ \sum_{t=0}^{\infty} \gamma^t R(s, a) \right]$.

\subsection{Policy Gradient Methods}
Policy gradient methods directly optimize the policy to maximize the expected return.
One of the key challenges in policy gradient methods is that aggressive updates can destabilize training.
To address this challenge, \citet{schulman2015trust} proposed trust region policy optimization (TRPO), which constrains the Kullback--Leibler (KL) divergence between the data-collection policy $\pi^{\text{old}}$ and the new policy $\pi^{\text{new}}$.
When the divergence remains under the trust region, TRPO theoretically guarantees monotonic improvement of the policy performance.
However, TRPO is computationally expensive because it requires second-order optimization.
To address this limitation, PPO~\citep{schulman2017proximal} was proposed as an efficient first-order alternative, using the following clipped surrogate objective $L^\text{PPO}$:
\begin{equation}
\label{eq:ppo_clip}
L^{\text{PPO}} = \hat{\mathbb{E}}_t \left[ \min \left(r_t\,\hat{A}_t, \, \text{clip}(r_t, 1 - \epsilon, 1 + \epsilon)\,\hat{A}_t \right) \right],
\end{equation}
where $r_t = {\pi^{\text{new}}(a_t|s_t)}/{\pi^{\text{old}}(a_t|s_t)}$ is the probability ratio between the new and old policies, $\hat{A}_t$ is an advantage estimate, and $\epsilon$ controls the clipping range.
Notably, the PPO clipping induces an implicit trust region by constraining the KL divergence on the order of $\epsilon^2$ (see Supplementary~\ref{sup_sec:PPO_clip_trust_region}).

\section{Divergence Misestimation in Mixture-of-Experts Policy Optimization}
\label{sec:divergence_misestimation}

The divergence of MoE policies can be misestimated.
Such misestimation is problematic for policy-gradient algorithms~\citep{schulman2015trust, schulman2017proximal} that regulate the divergence.
To formalize the misestimation, we consider a Gaussian policy whose mean is parameterized by a dense MoE.
Let $\mu_1, \dots, \mu_{N}$ denote the outputs of $N$ experts.
The MoE policy is then parameterized as
$\pi(a|s) = \mathcal{N}\!\left(m,\, \sigma^2 I\right) = \mathcal{N}\!\left(\sum_{e=1}^{N} w_e\, \mu_e,\; \sigma^2 I\right)$,
where $\mathbf{w} = [w_1, \dots, w_{N}]$ are obtained by applying a softmax to the router's output, ensuring their sum to one.
For simplicity, $\sigma$ is assumed to be fixed within each update step.

The measured KL divergence between the old and new policies is as follows:
\begin{align}
\text{KL}_\text{measured} = D(\pi^{\text{old}} \,\|\, \pi^{\text{new}}) = \|m^{\text{old}} - m^{\text{new}}\|^2 / 2\sigma^2,
\end{align}
where $m^{\text{old}} = \sum w_e^{\text{old}} \mu_e^{\text{old}}$ and $m^{\text{new}} = \sum w_e^{\text{new}} \mu_e^{\text{new}}$ denote the means of the old and new policies, respectively.
The difference $m^\text{old} - m^\text{new}$ yields the decomposition as follows:
\begin{align}
m^{\text{old}} - m^{\text{new}} = \underbrace{\sum w_e^{\text{old}} \left(\mu_e^{\text{old}} - \mu_e^{\text{new}}\right)}_{d_\mu} + \underbrace{\sum \left(w_e^{\text{old}} - w_e^{\text{new}}\right) \mu_e^{\text{new}}}_{d_w},
\end{align}
where $d_\mu$ and $d_w$ isolate the changes attributable to the experts and the router, respectively.
Then, the measured KL divergence can be expressed as follows:
\begin{align}
\text{KL}_\text{measured} = \frac{1}{2\sigma^2} \left(\|d_\mu\|^2 + 2\, d_\mu \cdot d_w + \|d_w\|^2\right).
\end{align}
We now show that the measured KL divergence can misestimate the true changes in each component.
To measure the true changes in each component, we decompose the policy update into two sequential steps that update the experts first and then the router.
Consider a partially updated policy, $\pi' = \mathcal{N}\left(\sum w_e^{\text{old}} \mu_e^{\text{new}}, \sigma^2 I\right)$, in which only the experts are updated while the router remains unchanged.
The fully updated policy $\pi^{\text{new}}$ is then obtained by updating only the router from $\pi'$.
The two-step path is as follows: $\pi^\text{old} \xrightarrow{\text{expert update}} \pi' \xrightarrow{\text{router update}} \pi^\text{new}$.

Then we can define the KL divergence for each update step in the path as follows:
\begin{align}
\text{KL}_{\text{expert}} = D(\pi^{\text{old}} \,\|\, \pi') = \frac{\|d_\mu\|^2}{2\sigma^2}, \qquad
\text{KL}_{\text{router}} = D(\pi' \,\|\, \pi^{\text{new}}) = \frac{\|d_w\|^2}{2\sigma^2}.
\end{align}
Therefore, the measured KL divergence can be rewritten as follows:
\begin{equation}
    \text{KL}_\text{measured} = \text{KL}_\text{router} + \text{KL}_\text{expert} + d_w \cdot d_\mu / \sigma^2.
\end{equation}
An analogous decomposition holds when the router is updated first, with $d_w$ and $d_\mu$ defined along that path.

\begin{definition}[Divergence Misestimation]
    \emph{Divergence misestimation} refers to the discrepancy between the measured divergence $\text{KL}_{\text{measured}}$ and the sum of component-wise divergences $\text{KL}_{\text{router}} + \text{KL}_{\text{expert}}$. 
    Specifically,
    \begin{align*}
        d_w \cdot d_\mu > 0 \;\;&\Longrightarrow\;\; \text{KL}_{\text{measured}} > \text{KL}_{\text{router}} + \text{KL}_{\text{expert}} \quad \text{(overestimation)}, \\
        d_w \cdot d_\mu < 0 \;\;&\Longrightarrow\;\; \text{KL}_{\text{measured}} < \text{KL}_{\text{router}} + \text{KL}_{\text{expert}} \quad \text{(underestimation)}.
    \end{align*}
\end{definition}

When overestimation occurs, the policy update is overly constrained by the divergence regularization, which can lead to conservative updates and suboptimal performance.
Underestimation has the opposite effect.
It makes the divergence regularization fail to constrain aggressive updates of each component, potentially leading to unstable training.
Notably, the sign of $d_w \cdot d_\mu$ is difficult to control explicitly during policy updates, as the gradient directions of the router and experts are not guaranteed to align or oppose each other.
Therefore, simple solutions such as adjusting the clipping range of PPO cannot effectively address the divergence misestimation.

\begin{figure}[t!]
    \centering
    \captionsetup{font=small}
    \begin{subfigure}{0.43\linewidth}
        \centering
        \includegraphics[width=\linewidth]{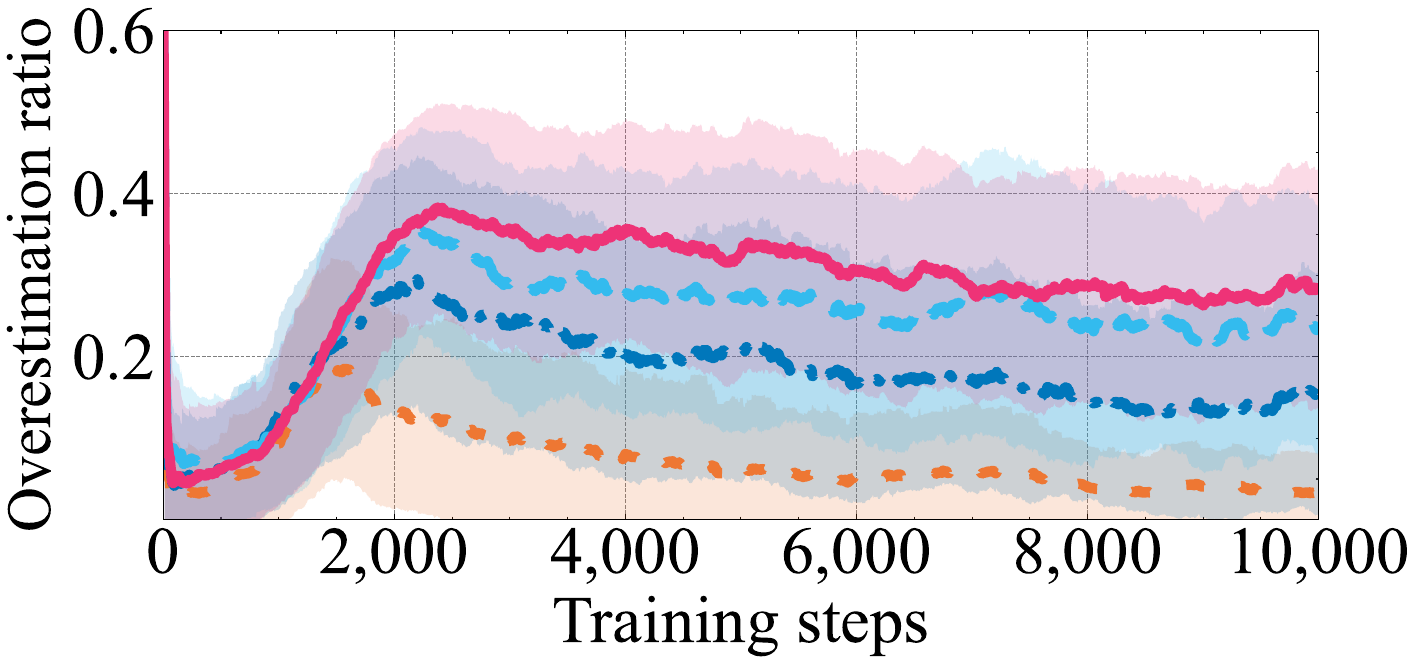}
        \vspace{-5mm}
        \caption{Overestimation}
        \label{fig:overest}
    \end{subfigure}
    \hfill
    \begin{subfigure}{0.43\linewidth}
        \centering
        \includegraphics[width=\linewidth]{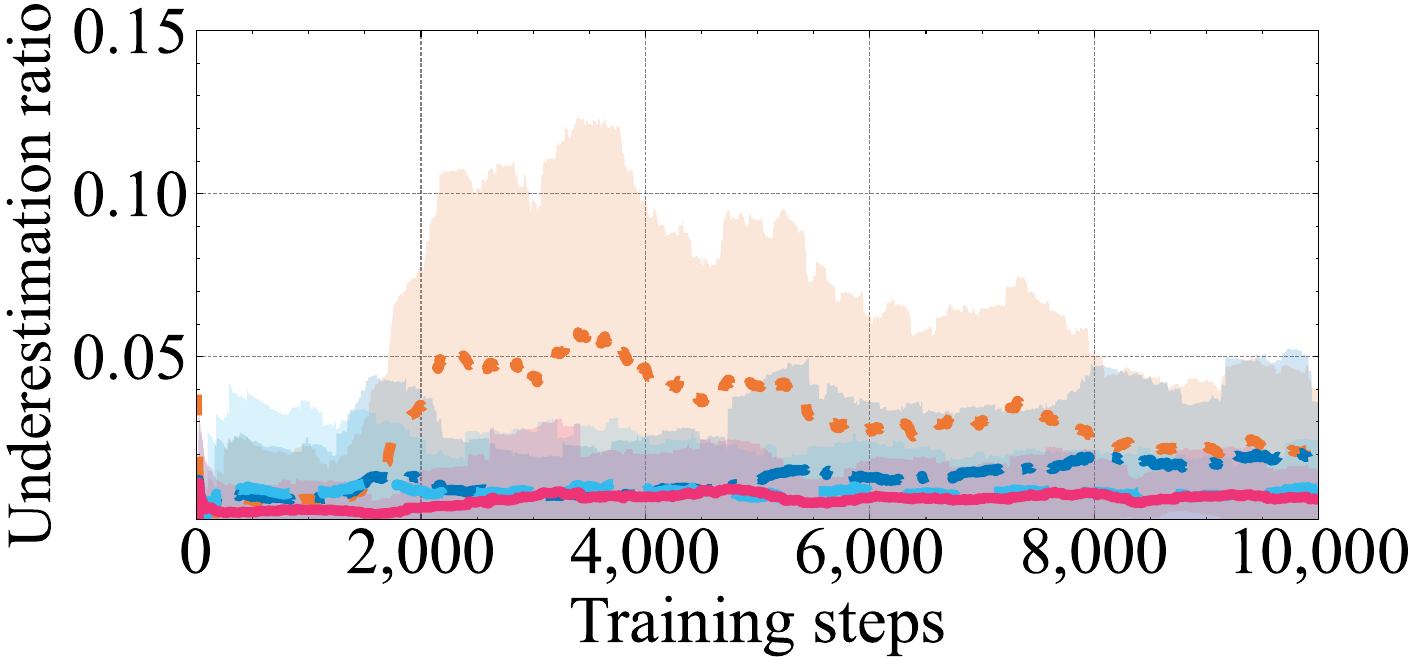}
        \vspace{-5mm}
        \caption{Underestimation}
        \label{fig:underest}
    \end{subfigure}
    \hfill
    \begin{minipage}{0.12\linewidth}
        \centering
        \vspace{-30mm}
        \includegraphics[width=\linewidth]{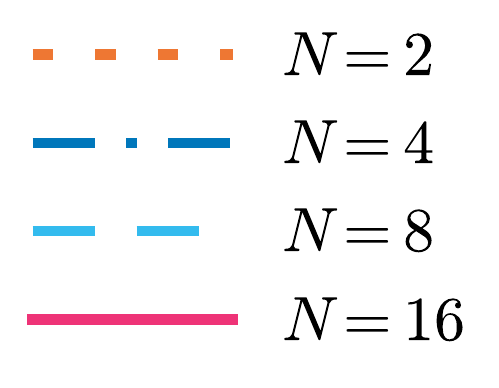}
    \end{minipage}
    \vspace{-2mm}
    \caption{
    Ratio of non-trivial policy divergence misestimation.
    $N$ is the number of experts.
    Overestimation becomes more frequent with more experts, whereas underestimation becomes more prevalent with fewer experts.
    Triviality threshold is $\epsilon^2$, where $\epsilon$ is the default PPO clipping ratio.
    }
    \label{fig:misestimation}
    \vspace{-6mm}
\end{figure}
Figure~\ref{fig:misestimation} shows the ratio of non-trivial divergence misestimation during the training of an MoE policy with vanilla policy-gradient algorithms.
A divergence misestimation is considered non-trivial when the difference between the measured KL divergence and the sum of component KL divergences exceeds the threshold. 
Because PPO typically constrains the KL divergence on the order of $\epsilon^2$, we set the threshold to $\epsilon^2$ with the default value of $\epsilon$.
Overestimation becomes more prevalent as the number of experts increases, while underestimation dominates with fewer experts.
Therefore, divergence misestimation is inherent to combining MoE with policy-gradient optimization; tuning the number of experts only switches its dominant mode between underestimation and overestimation.
Please refer to the Figure~\ref{fig:misest_dist} for the distribution of the misestimation and further analysis of its effect under training dynamics.

\section{Mixture of Experts for Universal Humanoid Motion Tracking}
\label{sec:mapo_algorithm}

In this study, we aim to train a humanoid robot to imitate diverse human motions~\citep{mahmood2019amass, harvey2020robust, li2023object}.
Following previous works~\citep{chen2025gmt, wang2026general, han2025kungfubot2, li2025clone}, we formulate this task as an RL problem and adopt an MoE architecture to enhance the expressiveness of the policy network.
To avoid the divergence misestimation issue, we propose MAPO\@, a policy optimization algorithm for MoE policies.

\subsection{MAPO: Mixture of Experts with Alternating Policy Optimization}
A simple solution to policy divergence misestimation is to store the mixing weights during data collection and use them to compute the probability ratio during policy optimization. 
However, in this case, the router is no longer part of the gradient computation and thus cannot be optimized with the RL objective. 
We thus face a trade-off between preventing divergence misestimation and optimizing the router with the RL objective.

To address this trade-off, we propose MAPO\@.
A key idea of MAPO is to alternate between expert and router updates, with the other component held fixed at its data-collection value.
The corresponding probability ratios for each update are
\begin{equation}
\label{eq:mapo_ratios}
r_t^{\text{expert}} = \frac{\mathcal{N}(\sum w_e^{\text{old}} \mu_e, \sigma^2 I)}{\mathcal{N}(\sum w_e^{\text{old}} \mu_e^{\text{old}}, \sigma^2 I)}, \quad
r_t^{\text{router}} = \frac{\mathcal{N}(\sum w_e \mu_e^{\text{old}}, \sigma^2 I)}{\mathcal{N}(\sum w_e^{\text{old}} \mu_e^{\text{old}}, \sigma^2 I)}.
\end{equation}
Following Equation~\ref{eq:ppo_clip}, the surrogate loss is then computed separately for each component using the clipped probability ratio, yielding gradients for the corresponding parameters.
Because each ratio can directly measure the amount of change in the corresponding component, the policy divergence misestimation can be prevented.
Additionally, because both components are used to compute the gradient, the router and experts can be optimized with the RL objective.
Notably, \emph{alternating} refers to separate gradient computations for the experts and the router, rather than an interleaved update schedule.
Both surrogate losses are computed from the same rollout data within a single update step.

\begin{figure}[t!]
    \centering
    \captionsetup{font=small}
    \includegraphics[width=1.0\linewidth]{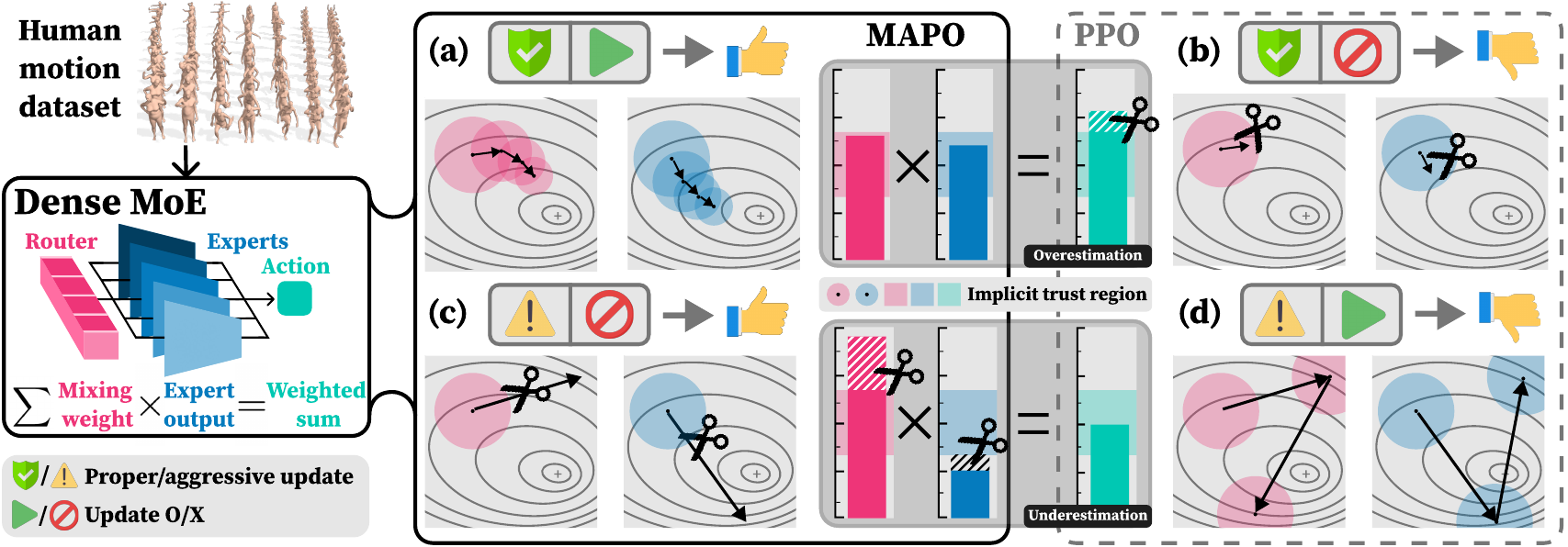}
    \vspace{-1mm}
    \caption{
Comparison of update mechanisms between MAPO and PPO\@.
Pink, blue, and blue-green indicate the router, experts, and mixed policy, respectively.
The bars in the center show the magnitudes of each component and the mixed policy.
The shaded regions show the implicit trust regions induced by clipping.
(a, c) MAPO measures the KL divergence of each component separately, allowing proper regulation of the updates.
(b, d) In contrast, because PPO measures the KL divergence of the mixed policy, the clipping mechanism may fail to regulate the updates properly.
}   
    \label{fig:architecture}
    \vspace{-6mm}
\end{figure}

Figure~\ref{fig:architecture} illustrates the differences in update mechanisms between MAPO and a standard policy gradient algorithm, PPO\@.
Specifically, PPO measures the KL divergence of the mixed policy, while MAPO measures each component separately for clipping.
When overestimation occurs, the measured divergence of the mixed policy can exceed the implicit trust region even though each component's update is admissible.
In this case, MAPO still updates both components properly, as shown in Figure~\ref{fig:architecture}~(a), whereas PPO blocks these admissible updates, as shown in Figure~\ref{fig:architecture}~(b).
When underestimation occurs, the mixed policy can stay within the trust region while individual components violate their own trust regions.
MAPO properly clips each component, as shown in Figure~\ref{fig:architecture}~(c), whereas PPO fails to clip the aggressive updates, as shown in Figure~\ref{fig:architecture}~(d).
Consequently, MAPO avoids the divergence misestimation in both cases, enabling more effective policy optimization.

Although MAPO avoids the misestimation issue, each update step optimizes a partially updated policy rather than the fully updated policy, introducing a gradient bias. 
For the expert parameters, the bias preserves the stationary points of the true gradient. 
For the router parameters, the gradient direction may deviate, but the bias depends only on the relative inter-expert deviations and vanishes when experts update coherently. 
While the bias persists when experts update incoherently, our experiments show that resolving the misestimation yields a favorable 
net effect.
See Supplementary~\ref{sup_sec:bias_analysis} for a detailed analysis.

\subsection{Teacher-Student Training Framework}
We adopt the teacher-student training framework~\citep{ji2025exbody2, he2025omnih2o, he2025hover, li2025amo}, in which a teacher is trained with privileged information and then distilled into a student for real-world deployment.

\textbf{Teacher Policy Training:\@}
The teacher policy is implemented as an MoE and trained with MAPO\@. 
For a fair comparison of the effect of MAPO, we closely follow prior humanoid motion tracking works~\citep{liao2025beyondmimic, luo2025sonic, he2025hover, ze2025twist2} for the observation space, reward function, and other training details~\citep{peng2018deepmimic}.
The policy outputs target joint positions, which a PD controller converts into joint torques.
The observation space consists of two parts. 
The first part, the proprioception state, contains the joint angles and velocities, root angular velocity, gravity vector, previous action, and per-link position, orientation, linear velocity, and angular velocity. 
The second part, the reference motion state, contains (i) future reference link positions and heading direction, (ii) current linear and angular velocities of each link, and (iii) per-link differences in current position and orientation between the robot and the reference.
The reward function consists of tracking and regularization terms. 
The tracking term encourages the robot to imitate the reference motion, while the regularization term penalizes undesirable behaviors such as joint position limit violations. 
See Table~\ref{tab:reward} for the full reward specification.

\textbf{Student Policy Distillation:\@}
Because some privileged observations, such as the linear velocity of the robot, are unavailable in the real world~\citep{he2025omnih2o}, the teacher policy cannot be directly deployed. 
Therefore, we distill it into a student policy that excludes privileged observations.
We use DAgger~\citep{ross2011reduction} for the distillation, where the student policy rolls out in the environment, and the teacher policy provides action labels for supervision.
Because the student is trained with simple supervised learning, we adopt a monolithic network for it.
The student policy's observation and domain randomization follow the settings from TWIST~\citep{ze2025twist}.

\section{Experiments}
\label{sec:experiments}
\subsection{Experiment Setup} 
We evaluate MAPO on a humanoid motion tracking task.
Because MAPO focuses on the training of MoE teacher policies, all evaluations in this section are conducted on the teacher policy, except for the real-world deployment.
We train both the teacher and student policies in the Isaac Lab simulator~\citep{mittal2025isaac}, with 4096 parallel environments on a single NVIDIA A6000 GPU\@.
The student policy is deployed on a real humanoid robot, Unitree G1 (29DoF)~\citep{unitree_g1}, for real-world demonstration.
We use AMASS~\citep{mahmood2019amass} and LAFAN1~\citep{harvey2020robust} datasets for training and OMOMO~\citep{li2023object} for evaluation.
All motions are retargeted to the humanoid using GMR~\citep{araujo2025retargeting} to account for morphological differences.
The MoE policy uses 6 experts, following prior works~\citep{chen2025gmt, han2025kungfubot2}.
For all methods, we use the standard PPO hyperparameters; the clipping parameter $\epsilon$ is set to 0.2, and the learning rate is adaptively adjusted following~\citet{rudin2022learning}. 
The other hyperparameters can be found in Table~\ref{tab:training_hyperparams}.

We compare MAPO against the following baselines. 
\textbf{PPO with MoE}: 
A dense MoE policy trained with PPO\@.
It directly captures how divergence misestimation affects MoE policy training.
\textbf{Monolithic PPO}:
A single monolithic policy trained with PPO\@.
It serves as a baseline to quantify the benefit of the MoE architecture.
\textbf{MAPO with fixed router}: 
A variant of MAPO that skips the router update, keeping the parameters of the router fixed.
Although divergence misestimation is reduced by fixing the router, end-to-end router optimization is prevented.
\textbf{HOVER}~\citep{he2025hover}: 
A universal motion tracker based on a teacher-student distillation framework. 
We re-implement their teacher policy on our humanoid platform and train it on the same dataset as our method.
Notably, we scale the penalty terms of HOVER to 20\% of the original values.
The penalty adjustment is the minimum required to enable training without unrealistic motion.
\textbf{TWIST~2}~\citep{ze2025twist2}:
A state-of-the-art humanoid motion tracking method.
We evaluate it using its released checkpoint.

\subsection{Experimental Results}

\paragraph{Tracking Performance:\@}
\begin{table}[t]
    \centering
    \captionsetup{font=small}
    \caption{
    Tracking metrics across methods.
    $E_\text{g-pos}$ and $E_\text{l-pos}$ are the global and root-relative (local) body position errors ($\mathrm{mm}$), $E_\text{acc}$ and $E_\text{vel}$ are the body acceleration ($\mathrm{mm/frame^2}$) and velocity ($\mathrm{mm/frame}$) errors.
    $E_\text{rt-h}$, $E_\text{rt-vel}$, $E_\text{rt-r}$, and $E_\text{rt-p}$ denote the root height ($\mathrm{m}$), linear velocity ($\mathrm{m/s}$), and roll and pitch errors (rad), respectively. 
    $N$ denotes the number of experts.
    The best performance for each metric is highlighted in \textbf{bold}, and the second best is \underline{underlined}.
    }
    \label{tab:tracking_metric}
    \vspace{2mm}
    {
    \scriptsize
    \setlength{\tabcolsep}{5.2pt}
    \begin{tabular}{c cl ccccccccc}
        \toprule
        \rowcolor{pink!30}
        \multicolumn{12}{l}{(a) Training set (AMASS~\citep{mahmood2019amass} + LAFAN1~\citep{harvey2020robust})} \\
        \midrule
        \multicolumn{3}{l}{Method} & Success $\uparrow$ & $E_\text{g-pos}\downarrow$ & $E_\text{l-pos}\downarrow$ & $E_\text{acc}\downarrow$ & $E_\text{vel}\downarrow$ & $E_\text{rt-h}\downarrow$ & $E_\text{rt-vel}\downarrow$ & $E_\text{rt-r}\downarrow$ & $E_\text{rt-p}\downarrow$\\ 
        \midrule
        \multicolumn{3}{l}{Monolithic PPO} & 81.78 \% & 165.7 & 96.50 & 5.787 & 16.35 & 0.060 & 0.532 & 0.169 & 0.136\\ 
        \multicolumn{3}{l}{PPO with MoE} & 83.38 \% & 153.1 & 87.98 & 5.499 & \underline{14.30} & 0.058 & 0.501 & \textbf{0.120} & \textbf{0.105}\\ 
        \multicolumn{3}{l}{MAPO with fixed router} & \underline{83.78 \%} & \underline{152.7} & \underline{86.65} & \underline{5.250} & 14.42 & \underline{0.052} & \underline{0.487} & 0.144 & 0.122\\ 
        \multicolumn{3}{l}{\textbf{MAPO (Ours)}} & \textbf{86.60 \%} & \textbf{147.4} & \textbf{78.10} & \textbf{4.576} & \textbf{11.78} & \textbf{0.048} & \textbf{0.406} & \underline{0.138} & \underline{0.117}\\
        \midrule
        \multirow{6}{*}{\rotatebox[origin=c]{90}{\shortstack[c]{Ablation on $N$}}}
                                & \multirow{2}{*}{$N=4$}  & PPO with MoE           & 83.96 \%          & 156.5            & 85.01            & 5.301            & 13.75            & 0.054            & 0.490            & 0.131            & 0.119            \\ 
                                &                         & \textbf{MAPO} & \textbf{86.87 \%}  & \textbf{143.6}   & \textbf{74.61}   & \textbf{4.672}   & \textbf{11.60}   & \textbf{0.050}   & \textbf{0.407}   & \textbf{0.119}   & \textbf{0.108}   \\ 
        \cmidrule(lr){2-3}
                                & \multirow{2}{*}{$N=8$}  & PPO with MoE           & 82.28 \%          & 164.4            & 93.47            & 5.841            & 15.51            & 0.064            & 0.538            & \textbf{0.126}   & \textbf{0.110}   \\ 
                                &                         & \textbf{MAPO} & \textbf{87.85 \%}  & \textbf{134.8}   & \textbf{71.65}   & \textbf{4.120}   & \textbf{10.06}   & \textbf{0.045}   & \textbf{0.360}   & 0.138            & 0.115            \\ 
        \cmidrule(lr){2-3}
                                & \multirow{2}{*}{$N=16$} & PPO with MoE           & 82.95 \%          & 159.6            & 94.08            & 6.175            & 15.54            & 0.061            & 0.528            & \textbf{0.118}   & \textbf{0.107}   \\ 
                                &                         & \textbf{MAPO} & \textbf{87.73 \%}  & \textbf{136.9}   & \textbf{72.58}   & \textbf{4.549}   & \textbf{11.02}   & \textbf{0.048}   & \textbf{0.375}   & 0.128            & 0.110            \\ 
        \midrule
        \rowcolor{blue!10}
        \multicolumn{12}{l}{(b) Test set (OMOMO~\citep{li2023object})} \\
        \midrule
        \multicolumn{3}{l}{Monolithic PPO} & 84.36 \% & 227.2 & \underline{86.43} & 6.014 & 12.44 & 0.035 & 0.356 & 0.157 & 0.134\\ 
        \multicolumn{3}{l}{PPO with MoE} & 82.78 \% & 241.4 & 93.90 & 6.451 & 13.77 & 0.041 & 0.397 & \underline{0.140} & 0.130\\ 
        \multicolumn{3}{l}{MAPO with fixed router} & \underline{85.21 \%} & \textbf{221.6} & 87.01 & 5.937 & 12.30 & \underline{0.031} & \underline{0.331} & 0.153 & 0.136\\ 
        \multicolumn{3}{l}{\textbf{MAPO (Ours)}} & \textbf{85.45 \%} & \underline{225.8} & \textbf{85.29} & \underline{5.764} & \textbf{11.76} & \textbf{0.030} & \textbf{0.326} & 0.151 & \underline{0.129}\\ 
        \midrule[0.2pt] 
        \multicolumn{3}{l}{HOVER~\citep{he2025hover}} & 83.39 \% & 272.2 & 122.81 & 6.001 & 12.46 & 0.058 & 0.382 & 0.193 & 0.163\\
        \multicolumn{3}{l}{TWIST~2~\citep{ze2025twist2}} & 57.85 \% & 386.6 & 104.72 & \textbf{4.661} & \underline{11.87} & 0.042 & 0.355 &  \textbf{0.080} & \textbf{0.074}\\ 
        \bottomrule
    \end{tabular}
    }
\end{table}

To evaluate the tracking performance, we measure the success rate and tracking errors in diverse metrics across all motions in the training and evaluation datasets, without any motion curation.
A motion is considered successfully tracked if the average position tracking error is lower than $0.5~ \mathrm{m}$ throughout the motion, following previous works~\cite{luo2023perpetual, he2025omnih2o}.
For fairness, we freeze the tracking error at its failure-time value after a failure, because excluding post-failure errors would underestimate the error of harder motions that fail early.
The results are summarized in Table~\ref{tab:tracking_metric}.
\textbf{MAPO} achieves the highest success rate and the lowest tracking errors across most metrics on both the training and evaluation datasets.
Additionally, we ablate the number of experts in the MoE policy.
As shown in Table~\ref{tab:tracking_metric}, \textbf{MAPO} consistently achieves better performance than \textbf{PPO~with~MoE} across all tested expert counts, except for the root roll and pitch errors.

\begin{figure}[b!]
    \centering
    \captionsetup{font=small}
    \begin{subfigure}{0.46\linewidth}
        \centering
        \includegraphics[width=\linewidth]{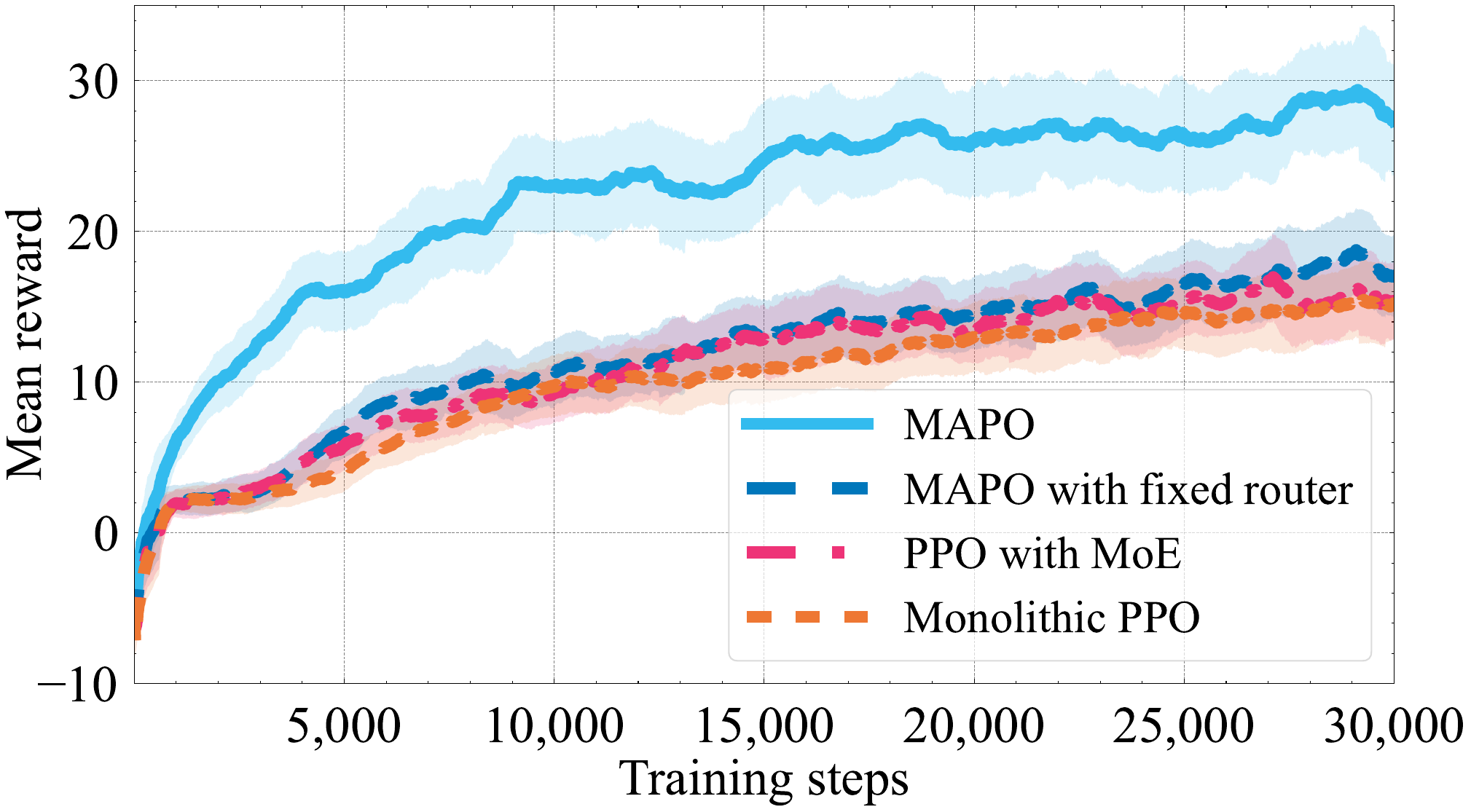}
        \vspace{-5mm}
        \caption{}
    \end{subfigure}
    \hfill
    \begin{subfigure}{0.21\linewidth}
        \centering
        \includegraphics[width=\linewidth]{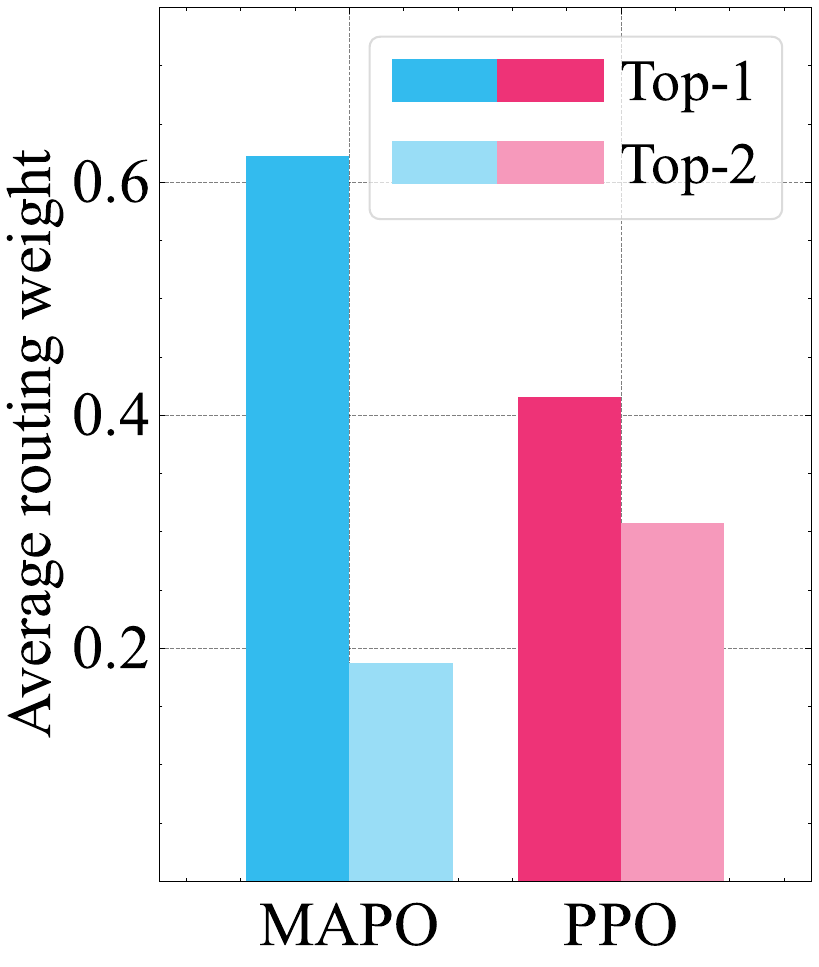}
        \vspace{-5mm}
        \caption{}
    \end{subfigure}
    \hfill
    \begin{subfigure}{0.3\linewidth}
        \centering
        \includegraphics[width=\linewidth]{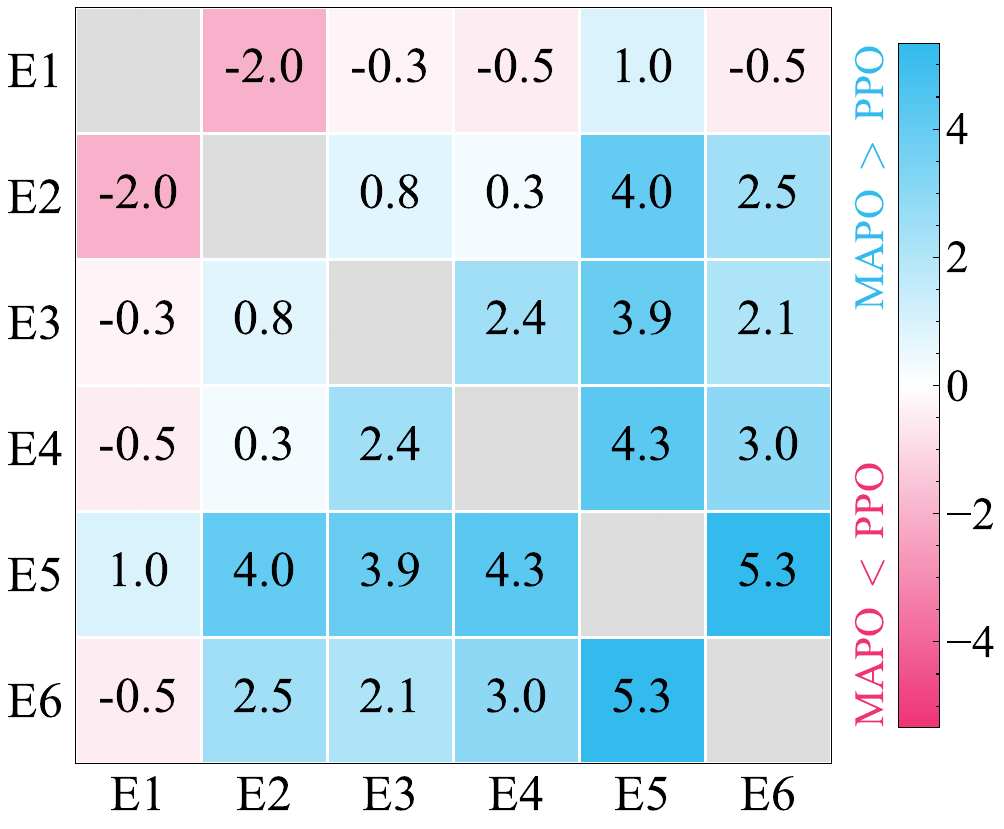}
        \vspace{-5mm}
        \caption{}
    \end{subfigure}
    \vspace{-2mm}
    \caption{
    (a) Reward learning curves. MAPO achieves substantially higher rewards than the baselines.
    (b) Sorted mixing weights of MAPO and PPO, averaged at each rank across inputs. 
    The misestimation of PPO prevents the mixing weights from sharpening.
    (c) Difference in pairwise expert output distances (MAPO $-$ PPO). 
    Positive entries indicate higher expert diversity in MAPO\@.
    }
    \label{fig:total_exp}
    \vspace{-5mm}
\end{figure}

\paragraph{Effectiveness of MAPO Training:\@}
Figure~\ref{fig:total_exp} (a) shows the effect of divergence misestimation on training performance.
\textbf{MAPO} achieves a significantly higher reward than \textbf{PPO with MoE}, demonstrating that avoiding misestimation substantially improves the training performance of MoE policies.
Furthermore, \textbf{MAPO with fixed router} achieves slightly higher reward than \textbf{PPO with MoE}, suggesting that the performance loss from misestimation is comparable to that from fixing the router.
Additionally, all MoE-based methods achieve higher reward than \textbf{Monolithic PPO}, confirming the effectiveness of the MoE architecture in humanoid motion tracking.

\paragraph{Expert Specialization:\@}

The specialization of dense MoE arises from differences in the mixing weights, which induce distinct gradients for each expert~\citep{jacobs1991adaptive}.
Therefore, sharper mixing weights are generally preferred, provided that all experts remain utilized across inputs.
However, although sharpened routing is desirable for specialization, it can exacerbate divergence overestimation: sharpening induces a large router-side change $d_w$, which can amplify the cross term $d_w \cdot d_\mu$.
Consequently, \textbf{PPO with MoE} tends to clip such updates, yielding less sharp routing than \textbf{MAPO} (Figure~\ref{fig:total_exp}~(b)).
The sharper routing of \textbf{MAPO} promotes better expert specialization, reflected in the generally larger pairwise expert output distances (Figure~\ref{fig:total_exp}~(c)). 
Each entry is the average L2 distance between the outputs of expert pairs on the same input, reported as (\textbf{MAPO} $-$ \textbf{PPO}).
Figure~\ref{fig:expert_activation} in the Supplementary shows per-expert activation across different motion types.

\begin{figure}[t!]
    \centering
    \captionsetup{font=small}
    \vspace{-5mm}
    \includegraphics[width=0.9\linewidth]{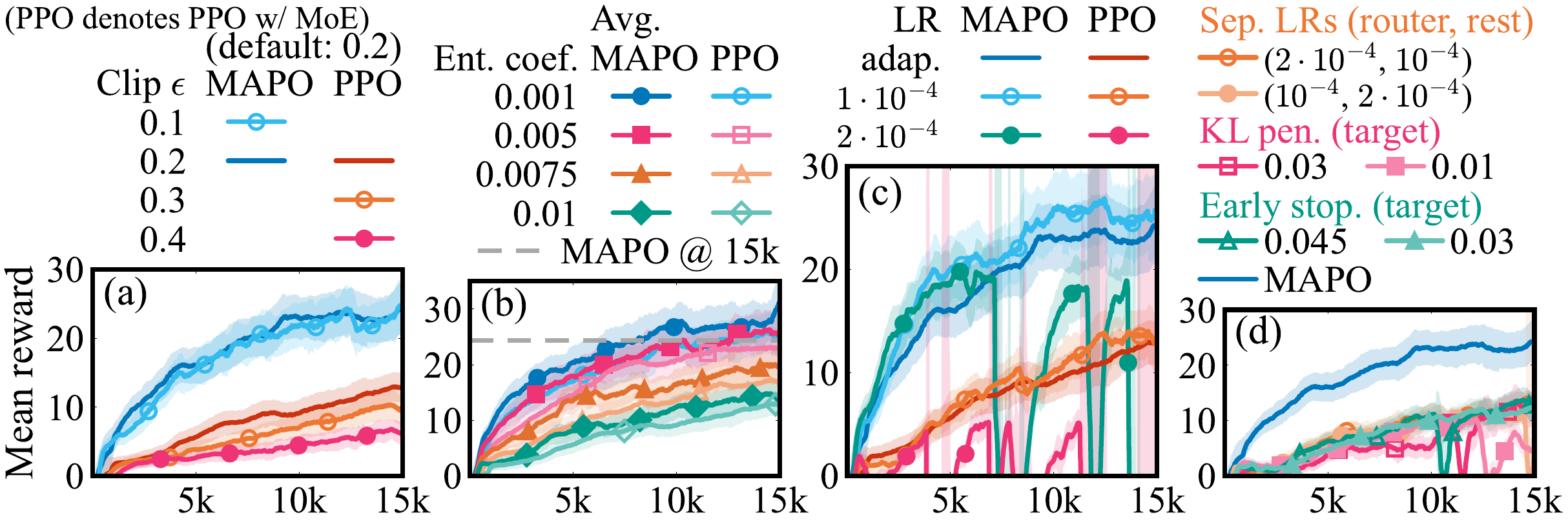}
    \caption{(a) Clipping parameter $\epsilon$ sweep. (b) Entropy coefficient sweep. (c) Fixed learning rate ablation. (d) Variants of PPO with MoE using a default learning rate of $10^{-4}$.}
    \label{fig:confounding_factor}
    \vspace{-5mm}
\end{figure}

\paragraph{Ablation of Confounding Factors:\@}
We examine potential confounding factors through the ablation studies in Figure~\ref{fig:confounding_factor}.
First, we test whether the performance gain of \textbf{MAPO} results from a larger update budget due to separate clipping of the expert and router updates (Figure~\ref{fig:confounding_factor}~(a)).
We scale the default clipping parameter $\epsilon$ by $0.5$ for \textbf{MAPO} and by $1.5$ and $2$ for \textbf{PPO with MoE}.
Increasing $\epsilon$ reduces the performance of \textbf{PPO with MoE}.
In contrast, \textbf{MAPO} with the smaller $\epsilon$ retains higher performance than \textbf{PPO with MoE} across all tested values.
The results suggest that the performance gain of \textbf{MAPO} does not arise solely from a larger update budget.

Second, the policy standard deviation $\sigma$ in \textbf{MAPO} is updated by both surrogate losses computed from the ratios in Equation~\ref{eq:mapo_ratios}.
However, the entropy bonus contributes to the update of $\sigma$ only once.
Therefore, the effective entropy coefficient for $\sigma$ can be halved.
To separate the effect of the reduced effective entropy coefficient from the effect of reduced divergence misestimation, we introduce \textbf{Averaged-MAPO}.
\textbf{Averaged-MAPO} scales each surrogate loss by $1/2$, thereby keeping the effective entropy coefficient for $\sigma$ unchanged.
Under matched entropy coefficients, \textbf{Averaged-MAPO} consistently achieves higher reward than \textbf{PPO with MoE} across all tested values (Figure~\ref{fig:confounding_factor}~(b)).

\paragraph{Consistency of Performance Gains:\@}
We examine whether the performance gain of \textbf{MAPO} persists under fixed learning rates.
We compare \textbf{MAPO} and \textbf{PPO with MoE} at learning rates of $10^{-4}$ and $2\times10^{-4}$ (Figure~\ref{fig:confounding_factor}~(c)).
\textbf{MAPO} consistently outperforms \textbf{PPO with MoE} at both learning rates.
Because training becomes unstable for both methods at $2\times10^{-4}$, we set $2\times10^{-4}$ as the maximum learning rate in the comparison.
Additionally, we compare \textbf{MAPO} with variants of \textbf{PPO with MoE} (Figure~\ref{fig:confounding_factor}~(d)).
We test separate learning rates for the router and remaining parameters.
We also evaluate alternative regularization techniques, including a KL penalty and early stopping.
However, the tested variants do not substantially improve on the default \textbf{PPO with MoE} and fail to match the performance of \textbf{MAPO}.

\paragraph{Qualitative Results:\@}
Figure~\ref{fig:title} visualizes representative motions where \textbf{MAPO} succeeds but \textbf{PPO~with~MoE} fails, under the same training setup.
The selected motions require complex whole-body coordination, including sitting in mid-air, abrupt full-body turns, deep crouching, and one-footed spinning. 
These motions are selected from 182 cases on which \textbf{MAPO} consistently succeeds while \textbf{PPO~with~MoE} consistently fails across 5 independent trials, whereas the reverse occurs for only 1 motion.
Notably, the five independent trials refer to evaluation rollouts of each trained policy.
Additionally, we demonstrate that the \textbf{MAPO}-trained policy can serve as an effective teacher for distillation. 
Figure~\ref{fig:title} also shows a real-world deployment of the student policy. 
The robot successfully tracks a wide range of reference motions, including challenging ones that require precise balance, such as squatting and bowing.
Videos can be found on our project website.

\section{Conclusion}
\label{sec:conclusion}
In this study, we identified the divergence misestimation phenomenon that arises when MoE policies are trained with policy-gradient algorithms.
To avoid the misestimation, we proposed MAPO, which alternately optimizes the router and experts, enabling more effective training and better tracking performance for humanoid whole-body control.
The resulting policy performs complex whole-body motions well and is distilled into a deployable student policy on a real humanoid robot.

\section{Limitations}
\label{sec:limitations}
While MAPO enables effective training of MoE policies with policy-gradient algorithms, further encouraging expert specialization remains an open question. 
One potential direction is to introduce an auxiliary loss that promotes discriminability between expert actions, inspired by prior works on unsupervised skill discovery~\cite{eysenbach2019diversity, park2024metra, strouse2022learning, hu2024disentangled, cathomen2025divide}.
Another limitation is that our analysis assumes dense softmax routing with $\sigma$ fixed within each update step.
The analysis does not cover learned per-expert variances or top-$k$ sparse routing.
Nevertheless, divergence misestimation may also occur in these settings.

\clearpage


\bibliography{references}  

\clearpage

\section*{Supplementary Material}
\appendix
\numberwithin{table}{section}
\numberwithin{figure}{section}
\section{Implicit Trust Region from clipping}
\label{sup_sec:PPO_clip_trust_region}
Let $r_t = 1 + \delta \quad (|\delta| \leq \epsilon)$, then the KL divergence between the new and old policies can be expressed as,
\begin{equation}
D(\pi^{\text{old}} || \pi^{\text{new}}) = \mathbb{E} \left[ \log \frac{1}{r_t} \right] = \mathbb{E} \left[ -\log (1 + \delta) \right].
\end{equation}

Because $\log (1 + \delta) \approx \delta - \delta^2/2$ for small $\delta$, we have,
\begin{equation}
D(\pi^{\text{old}} || \pi^{\text{new}}) \approx \mathbb{E} \left[ -\delta + \delta^2/2 \right].
\end{equation}

Because $\mathbb{E}_{a \sim \pi^\text{old}(\cdot|s)}[\delta] = \int \pi_\Theta(a|s)\, da - 1 = 0$, the KL divergence can be approximated as follows:
\begin{equation}
D(\pi^{\text{old}} || \pi^{\text{new}}) \approx \mathbb{E} \left[\delta^2/2 \right] = O(\delta^2) \leq O(\epsilon^2).
\end{equation}

It means the clipping of PPO implicitly constructs a trust region by constraining the KL divergence to $O(\epsilon^2)$.

\section{Bias Analysis of MAPO}
\label{sup_sec:bias_analysis}
Recall that the dense MoE policy can be modeled as $\pi_\Theta(a|s) = \mathcal{N}(m_\Theta, \sigma^2 I)$, where $m_\Theta = \sum_e w_e \mu_e$ and $\Theta=\left\{ \psi, \theta_{1:N} \right\}$.
The mixing weights $\mathbf{w} = [w_1, w_2, \dots, w_N]$ are obtained by applying a softmax to the output of the router.
The $e$-th expert outputs the deterministic mean $\mu_e = \mu_e(s; \theta_e)$.
For simplicity, we assume that all experts share the same variance $\sigma^2$.

\subsection{Score Function Decomposition}
Consider the score function decomposition for the policy gradient algorithm.
Define $h(s, a)$ as, 
\begin{equation}
    \nabla_m \log \pi(a|s) = \frac{a - m}{\sigma^2} \doteq h(s, a).
\end{equation}
Then, 
\begin{align}
    \nabla_\Theta \log \pi_\Theta(a|s) &= \nabla_m \log \pi(a|s) \, \nabla_\Theta m_\Theta \\
    &= h(s, a) \nabla_\Theta m_\Theta.
\end{align}
The $\nabla_\Theta$ consists of gradients for the parameters of the router and the experts, which can be expressed as,
\begin{equation}
\nabla_\psi m_\Theta = \sum_e \frac{\partial w_e}{\partial \psi} \mu_e, \text{ and }
\nabla_{\theta_k} m_\Theta = w_k \frac{\partial \mu_k}{\partial \theta_k}.
\end{equation}
Therefore,
\begin{equation}
    \nabla_\psi \log \pi_\Theta (a|s) = h \sum_e \frac{\partial w_e}{\partial \psi} \mu_e, \text{ and }
    \nabla_{\theta_k} \log \pi_\Theta (a|s) = h w_k \frac{\partial \mu_k}{\partial \theta_k}.
\end{equation}

\subsection{PPO Gradient}

Because the gradient of PPO when it is clipped is zero, we only consider the case when it is not clipped.
In this case, the PPO gradient is identical to the gradient of true importance sampling. 

Recall, 
\begin{equation}
L^{\text{PPO}} = \hat{\mathbb{E}} \left[ \min \left( r(s, a) \hat{A}, \text{clip}(r(s, a), 1 - \epsilon, 1 + \epsilon) \hat{A} \right) \right]
\end{equation}
where $r(s, a) = {\pi_\Theta(a|s)}/{\pi^{\text{old}}(a|s)}$. 

Notably,
\begin{equation}
\nabla_\Theta L^{\text{PPO}} = \hat{\mathbb{E}} \left[ \nabla_\Theta r(s, a) \hat{A} \right] = \hat{\mathbb{E}} \left[ r(s, a) h(s, a) \nabla_\Theta m(s ; \Theta) \hat{A} \right]
\end{equation}

\subsection{MAPO Gradient}

Let, $m_1(s; \psi) = \sum_e w_e(s, \psi) \mu_e^\text{old}(s)$ and $m_2(s; \theta_{1:N}) = \sum_e w_e^\text{old}(s) \mu_e(s; \theta_{1:N})$.
Then, we can define the following quantities,
\begin{align}
    \pi_1 &= \mathcal{N}(m_1, \sigma^2 I), \quad r_1 = {\pi_1}/{\pi^\text{old}}, \quad h_1 = {(a - m_1)}/{\sigma^2} \\
    \pi_2 &= \mathcal{N}(m_2, \sigma^2 I), \quad r_2 = {\pi_2}/{\pi^\text{old}}, \quad h_2 = {(a - m_2)}/{\sigma^2}.
\end{align}

Similar to the PPO case, we only consider the case when the MAPO gradient is not clipped.
Then, the MAPO loss function can be expressed as,

\begin{equation}
    L^{\text{MAPO}} = \hat{\mathbb{E}} \left[ r_1(s, a) \hat{A} \right] + \hat{\mathbb{E}} \left[ r_2(s, a) \hat{A} \right].
\end{equation}

Then the gradient of MAPO with respect to each parameter can be expressed as,
\begin{align}
\nabla_\psi L^\text{MAPO} &= \hat{\mathbb{E}} \left[ r_1 h_1(s, a) \nabla_\psi m_1(s; \psi) \hat{A} \right] \\
\nabla_{\theta_k} L^\text{MAPO} &= \hat{\mathbb{E}} \left[ r_2 h_2(s, a) \nabla_{\theta_k} m_2(s; \theta_{1:N}) \hat{A} \right],
\end{align}
where $\nabla_\psi m_1(s; \psi) = \sum_e \frac{\partial w_e}{\partial \psi} \mu_e^\text{old}$ and $\nabla_{\theta_k} m_2(s; \theta_{1:N}) = w_k^\text{old} \frac{\partial \mu_k}{\partial \theta_k}$.

\subsection{Effect of Clipping}
Because the clipping implicitly constrains the KL divergence induced by each of the two intermediate ratios $r_1, r_2$ to $O(\epsilon^2)$, it yields the following inequalities:

\begin{align}
    \frac{||m_1 - m_\text{old}||}{\sigma} &= \frac{1}{\sigma} ||\sum (w_e - w_e^\text{old}) \mu_e^\text{old}|| \leq \epsilon \\
    \frac{||m_2 - m_\text{old}||}{\sigma} &= \frac{1}{\sigma} ||\sum (\mu_e - \mu_e^\text{old}) w_e^\text{old}|| \leq \epsilon.
    \label{eq:clip_m2}
\end{align}

\subsection{Bias Analysis}
There are two sources of bias in the MAPO gradient.
The first is the bias in the mean gradient term,~$\nabla m$.
The second is the bias in the importance sampling ratio, $r$, and score factor $h$.
As we will show, the second source is of the same order as the surrogate approximation error of PPO and does not change the qualitative picture; we focus first on the bias in the mean gradient term.

\paragraph{Mean Gradient Bias in Expert Parameters:\@}

We have,
\begin{align}
    \nabla_{\theta_k} m_\Theta &= w_k \frac{\partial \mu_k}{\partial \theta_k} \quad {\text{(for PPO)}} \\
    \nabla_{\theta_k} m_2 &= w_k^\text{old} \frac{\partial \mu_k}{\partial \theta_k} \quad {\text{(for MAPO)}}.
\end{align}

Therefore, the MAPO gradient on each expert block is the true importance sampling gradient on that block scaled by a per-expert positive factor $w_k^\text{old} / w_k$.
As these scalars differ across experts, the aggregate gradient direction in $\theta_{1:N}$-space is not preserved in general; however, each per-expert block vanishes exactly when its true counterpart does, therefore the stationary points of MAPO coincide with those of the true importance sampling gradient.

\paragraph{Mean Gradient Bias in Router Parameters:\@}
The bias in the mean gradient term for the router parameters can be expressed as,
\begin{equation}
b_\psi^\text{mean} = \sum_e \frac{\partial w_e}{\partial \psi} (\mu_e - \mu_e^\text{old}).
\end{equation}

Notably, unlike the expert parameters, the MAPO gradient with respect to the router parameters can have a directional bias.
The following cancellation shows that this bias depends only on the \emph{relative} deviations between experts, not on their absolute magnitudes.

Because the sum of the mixing weights is 1, we have,
\begin{equation}
\sum_e \frac{\partial w_e}{\partial \psi} = \frac{\partial}{\partial \psi} \sum_e w_e = 0.
\end{equation}

Then, for any constant $\bar{\xi}$ independent of the expert index $e$, we have
\begin{equation}
\sum_e \frac{\partial w_e}{\partial \psi} (\mu_e - \mu_e^\text{old}) = \sum_e \frac{\partial w_e}{\partial \psi} \left[(\mu_e - \mu_e^\text{old}) - \bar{\xi}\right], \qquad \text{because } \sum_e \frac{\partial w_e}{\partial \psi} \bar{\xi} = \bar{\xi} \sum_e \frac{\partial w_e}{\partial \psi} = 0.
\end{equation}

We define
\begin{equation}
\xi_e = (\mu_e - \mu_e^\text{old}), \text{ and } \bar{\xi} = \frac{1}{N} \sum_e \xi_e = \frac{1}{N} \sum_e (\mu_e - \mu_e^\text{old}),
\end{equation}
then, we have
\begin{equation}
\label{eq:router_bias_centered}
b_\psi^\text{mean} = \sum_e \frac{\partial w_e}{\partial \psi} (\xi_e - \bar{\xi}).
\end{equation}
Therefore, the bias depends on the relative difference between each expert, rather than the absolute value of the difference.
In particular, when all the experts are updated coherently, $\xi_e - \bar{\xi} \approx 0$ and the bias vanishes.

\paragraph{Bound on the Rotation Angle on the Router Gradient:\@}
We now show that the directional discrepancy between the MAPO router gradient $\hat g := \nabla_{\psi} m_1$ and the true mean gradient $g := \nabla_{\psi} m_{\Theta}$ is also controlled by the inter-expert variance.
Within a single update step, the trust region in $\psi$ keeps the router parameters close to their old values, and the router Jacobian is approximately unchanged: $\partial w_e/\partial \psi \approx \partial w_e^{\mathrm{old}}/\partial \psi$.
Under this approximation, the gradient discrepancy reduces to the mean-gradient bias,
\begin{equation}
    g - \hat g \;\approx\; b_{\psi}^{\mathrm{mean}}.
\end{equation}
Decomposing $b_{\psi}^{\mathrm{mean}}$ into components parallel and perpendicular
to $\hat g$, the angle $\theta$ between $g$ and $\hat g$ satisfies
\begin{equation}
    \sin\theta \;\lesssim\; \frac{\lVert b_{\psi}^{\mathrm{mean}} \rVert}{\lVert \hat g \rVert},
\end{equation}
where the $\lesssim$ absorbs only the first-order Jacobian approximation above.

Applying the Cauchy--Schwarz inequality to~(\ref{eq:router_bias_centered}),
\begin{equation}
    \lVert b_{\psi}^{\mathrm{mean}} \rVert
    \;\le\;
    \sqrt{\sum_{e} \Bigl\lVert \tfrac{\partial w_e}{\partial \psi} \Bigr\rVert^{2}}
    \;\cdot\;
    \sqrt{\sum_{e} \lVert \xi_e - \bar{\xi} \rVert^{2}}.
\end{equation}
\begin{wrapfigure}{r}{0.32\textwidth}
    \vspace{-\intextsep}
    \centering
    \captionsetup{font=small}
    \includegraphics[width=\linewidth]{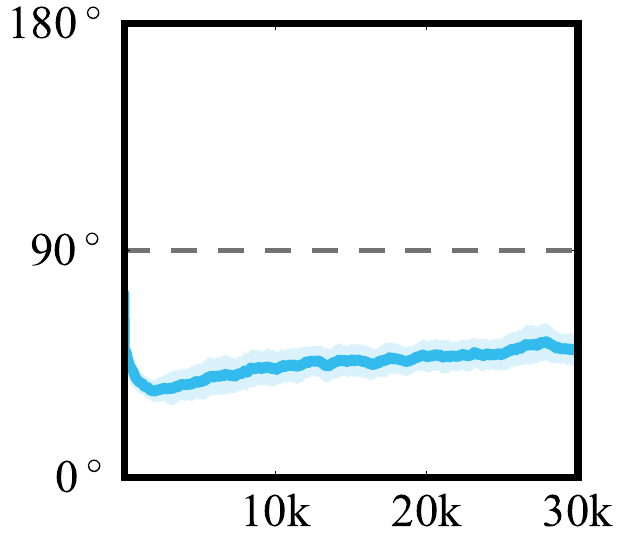}
    \vspace{-6mm}
    \caption{
    Router-gradient angle.
    }   
    \label{fig:bias_angle}
    \vspace{-\intextsep}
\end{wrapfigure}
Because $\hat g = \sum_e (\partial w_e^{\mathrm{old}}/\partial \psi)\, \mu_e^{\mathrm{old}}$
shares the same router Jacobian as the first factor above, that factor cancels
to leading order in the ratio $\lVert b_\psi^{\mathrm{mean}} \rVert / \lVert \hat g \rVert$.
The rotation angle is therefore governed primarily by the inter-expert variance
$\sum_e \lVert \xi_e - \bar{\xi} \rVert^{2}$, not by the absolute expert deviations
or the router sensitivity.
Whether this variance remains small in practice depends on the training
dynamics and is not guaranteed by the trust-region constraint
$\lVert \sum_e w_e^{\mathrm{old}} \xi_e \rVert \le \epsilon \sigma$ from Equation~\ref{eq:clip_m2}
alone.
Figure~\ref{fig:bias_angle} shows the changes in the router-gradient angle during training.
The router-gradient angle remains below $90\,^{\circ}$ throughout training.

\textbf{Difference in Importance Sampling Ratio and Score Factor:\@}
Because of the clipping, the deviations in the importance sampling ratio and score factor are of order $O(\epsilon)$,
\begin{equation}
|r_i - r| \approx O(\epsilon), \text{ and } |h_i - h| = |m_i - m|/\sigma^2 \approx O(\epsilon) \quad (i = 1, 2).
\end{equation}

Decomposing the gradient difference as
\[
r_i h_i \nabla m_i - r h \nabla m = r h \, (\nabla m_i - \nabla m) + (r_i h_i - r h) \, \nabla m_i,
\]
the first term is the mean-gradient bias analyzed above, while the second contributes an additional $O(\epsilon)$ correction because $r_i h_i - r h = O(\epsilon)$. 
Importantly, within the trust region $r \approx 1$, therefore the leading $r h$ pre-factor is itself close to the score $h$, and the residual error from the $(r_i, h_i)$ deviation is on the same order as PPO's surrogate approximation error and does not alter the qualitative conclusions of the mean-gradient analysis.

\section{Training Details}
\subsection{Reward Function}
\begin{table}[t!]
\centering
\caption{Reward functions for teacher training.
Superscripts $p$ and $g$ denote proprioception and goal states, respectively.
$\mathbf{o}$, $\mathbf{p}$, $\mathbf{v}$, $\boldsymbol{\omega}$, $\mathbf{a}$, $\mathbf{q}$, and $\mathbf{F}$ are orientation, position, linear velocity, angular velocity, policy action, joint position, and contact force.
Subscripts specify the reference point: \emph{root} for the pelvis link, \emph{body} for tracked body links in $\mathcal{B}$, \emph{rel-body} for body link quantities expressed relative to the root frame, and \emph{last} for the previous timestep.
$\mathbb{I}[\cdot]$ is the indicator function.}
    \vspace{2mm}
    \label{tab:reward}
    \begin{tabular}{llc}
        \toprule
        \textbf{Reward term} & \textbf{Equation} & \textbf{Weight} \\
        \midrule
        Root orientation & $\text{exp}\left(- ||\mathbf{o}^p_\text{root} - \mathbf{o}^g_\text{root}||^2_2  / 0.4^2 \right)$ &  1.0 \\
        Relative body-link position & $\text{exp}\left(- \frac{1}{|\mathcal{B}|} \sum_{b \in \mathcal{B}} ||\mathbf{p}^p_{\text{rel-body},b} - \mathbf{p}^g_{\text{rel-body},b}||^2_2/ 0.05^2 \right)$ & 20.0 \\
        Relative body-link orientation & $\text{exp}\left(- \frac{1}{|\mathcal{B}|} \sum_{b \in \mathcal{B}} ||\mathbf{o}^p_{\text{rel-body},b} - \mathbf{o}^g_{\text{rel-body},b}||^2_2  / 0.4^2 \right)$ & 1.0 \\
        Body-link linear velocity & $\text{exp}\left(- \frac{1}{|\mathcal{B}|} \sum_{b \in \mathcal{B}} ||\mathbf{v}^p_{\text{body},b} - \mathbf{v}^g_{\text{body},b}||^2_2 / 1.0 \right)$ & 1.0 \\
        Body-link angular velocity & $\text{exp}\left(- \frac{1}{|\mathcal{B}|} \sum_{b \in \mathcal{B}} ||\boldsymbol{\omega}^p_{\text{body},b} - \boldsymbol{\omega}^g_{\text{body},b}||^2_2 / 3.14^2 \right)$ & 1.0 \\
        \midrule
        Action rate & $||\mathbf{a} - \mathbf{a}_\text{last}||^2_2$ & -0.2 \\
        Joint limit & $\sum_j \mathbb{I} \left[ \mathbf{q}_j \notin [\mathbf{q}^\text{min}_j, \mathbf{q}^\text{max}_j] \right]$ & -10.0  \\
        Undesired contact & $\sum_{c \notin \left\{ \text{ankles, wrists} \right\}} \mathbb{I} \left[ ||\mathbf{F}_c|| > 1.0 N \right] $ & -0.1 \\
        \bottomrule
    \end{tabular}
\end{table}

\begin{figure}[b!]
    \centering
    \captionsetup{font=small}
    \includegraphics[width=1.0\linewidth]{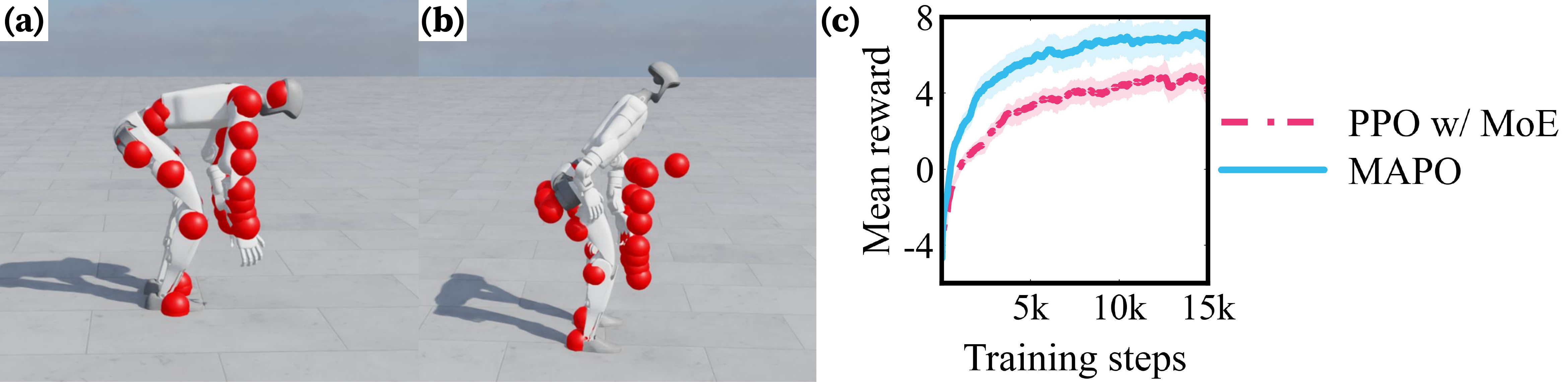}
    \caption{
    Comparison of reward settings and training performance.
    (a) The reward setting used in this work.
    (b) The reward setting of BeyondMimic~\citep{liao2025beyondmimic}.
    The reward settings differ mainly in the scale and sharpness of the relative body-link position reward.
    (c) Reward learning curves of MAPO and PPO with MoE under the BeyondMimic reward setting.
    }
    \label{fig:reward_compare}
    \vspace{-5mm}
\end{figure}
The reward function for the teacher training is almost similar to the prior works~\cite{liao2025beyondmimic,luo2025sonic}.
The specific reward functions are listed in Table~\ref{tab:reward}.
We find that the sharper relative body link position reward encourages the robot to fully complete the reference motion, particularly for deep squatting and bowing. 
Without it, the robot tends to follow the reference motion only partially, stopping halfway through such motions, as shown in Figure~\ref{fig:reward_compare}~(b).
Notably, \textbf{MAPO} also achieves higher reward than \textbf{PPO with MoE} under the BeyondMimic reward setting (Figure~\ref{fig:reward_compare}~(c)).

\subsection{Training Hyperparameters}
\begin{table}[t!]
    \centering
    \captionsetup{font=small}
    \caption{Teacher policy training hyperparameters.}
    \vspace{2mm}
    \label{tab:training_hyperparams}
    \begin{tabular}{lc}
        \toprule
        \textbf{Hyperparameter} & \textbf{Value} \\
        \midrule
        Num parallel environments & 4096 \\
        Num steps per environment & 24 \\
        Num mini-batches & 4 \\
        Learning epochs & 5 \\
        \midrule
        Discount $\gamma$ & 0.99 \\
        GAE $\lambda$ & 0.95 \\
        Clip parameter $\epsilon$ & 0.2 \\
        Entropy coefficient & 0.01 \\
        Learning rate & Adaptive \\
        Max grad norm & 1.0 \\
        \midrule
        Init noise std & 1.0 \\
        Shared actor encoder hidden layers & [512, 256] \\
        Shared encoder final feature dimension & 256 \\ 
        Router hidden layers & [256, 128] \\
        Expert hidden layers & [128, 128] \\
        Value function hidden layers & [512, 256, 128] \\
        \bottomrule
    \end{tabular}
\end{table}

The training hyperparameters for the teacher policy are listed in Table~\ref{tab:training_hyperparams}.
The Adam~\citep{kingma2015adam} optimizer is used for optimization, and exponential linear units (ELUs)~\citep{clevert2016fast} are used as the activation function in all networks.
Additionally, the teacher uses a shared encoder that extracts features from the observation, which are fed into both the router and experts~\citep{lee2015m, d2020sharing}.
Notably, for the alternating optimization, we detach the gradients from the router to the shared encoder.
For the student policy distillation, a fixed learning rate of $2\times10^{-4}$ is used, and the student policy network adopts hidden layers of sizes $[512, 256, 128]$.

\subsection{Domain Randomization}
\begin{table}[b!]
    \centering
    \captionsetup{font=small}
    \caption{Domain randomization parameters. $\mathcal{U}(a, b)$ denotes a uniform distribution over the interval $[a, b]$.}
    \vspace{2mm}
    \label{tab:domain_randomization}
    \begin{tabular}{lc}
        \toprule
        \textbf{Domain Randomization Parameter} & \textbf{Value} \\
        \midrule
        Joint P gain scale & $\mathcal{U}(0.75, 1.25) \times \text{default}$ \\
        Joint D gain scale & $\mathcal{U}(0.75, 1.25) \times \text{default}$ \\
        Root mass & $\mathcal{U}(0.75, 1.25) $ + $ \text{default}$ kg \\
        Root center of mass offset (x, y, z) & $\mathcal{U}([-0.1, 0.1], [-0.1, 0.1], [-0.1, 0.1])$ m \\
        Control delay & $\mathcal{U}(0, 60)$ ms \\
        \midrule
        Push robot magnitude & $\mathcal{U}(-1.0, 1.0)$ m/s \\
        Push robot interval & 5 s \\
        \midrule
        Reference motion offset (x, y, z) & $\mathcal{U}([-0.02, 0.02], [-0.02, 0.02], [-0.1, 0.1])$ m \\
        \bottomrule
    \end{tabular}
\end{table}
 
To improve the robustness of the trained policy, we apply domain randomization during training.
Details of the randomized parameters are listed in Table~\ref{tab:domain_randomization}.

\section{Additional Experimental Results}
\paragraph{Per-Expert Activation:\@}
\begin{figure}[t!]
    \centering
    \captionsetup{font=small}
    \vspace{-3mm}
    \includegraphics[width=0.8\linewidth]{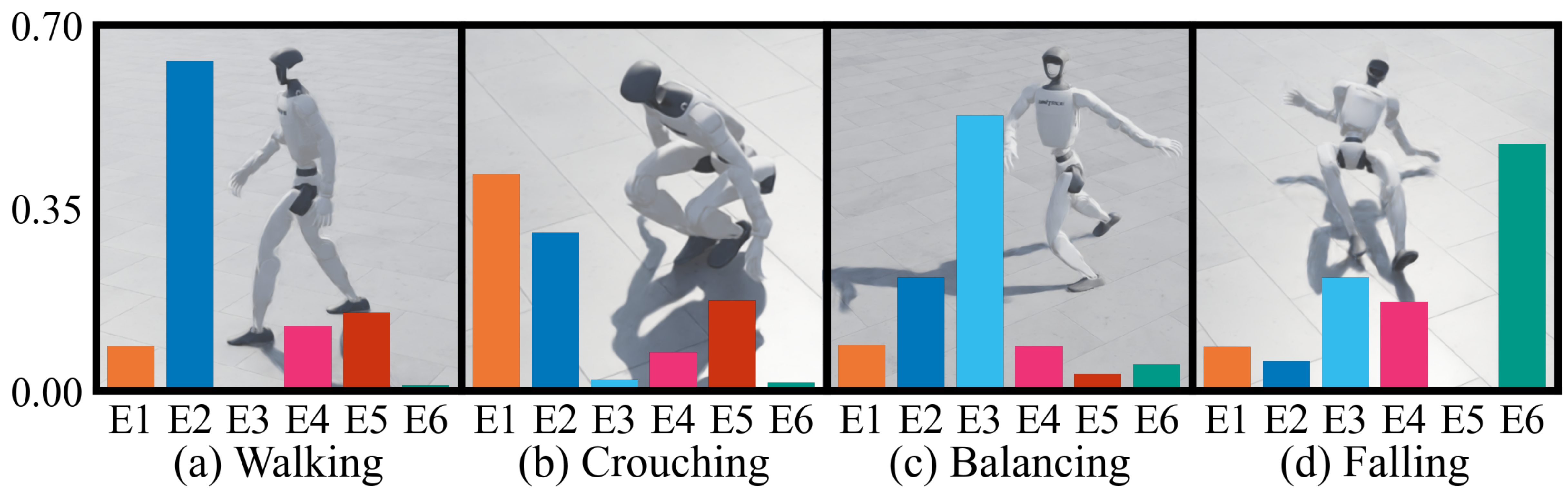}
    \caption{
    Per-expert activation according to the motion type.
    }
    \label{fig:expert_activation}
    \vspace{-7mm}
\end{figure}
Figure~\ref{fig:expert_activation} shows per-expert activation across different motion types.
The router assigns higher mixing weights to different experts depending on the motion type, indicating expert specialization.
Expert 2 (E2) is primarily activated during walking and upper-body motions while standing (Figure~\ref{fig:expert_activation}~(a)).
E4 and E5 support E2 during different phases of walking.
Although E1 appears redundant in Figure~\ref{fig:total_exp}~(c), E1 is activated for motions that require a squatting posture (Figure~\ref{fig:expert_activation}~(b)).
E3 and E6 are activated during balance recovery and unavoidable falls, respectively (Figure~\ref{fig:expert_activation}~(c, d)).
Because we do not manually curate the dataset, some reference motions may be infeasible for the robot.
Concentrating activation during unavoidable falls in E6 may reduce interference from infeasible motions during the training of the remaining experts.

\begin{figure}[b!]
    \centering
    \captionsetup{font=small}
    \includegraphics[width=1.0\linewidth]{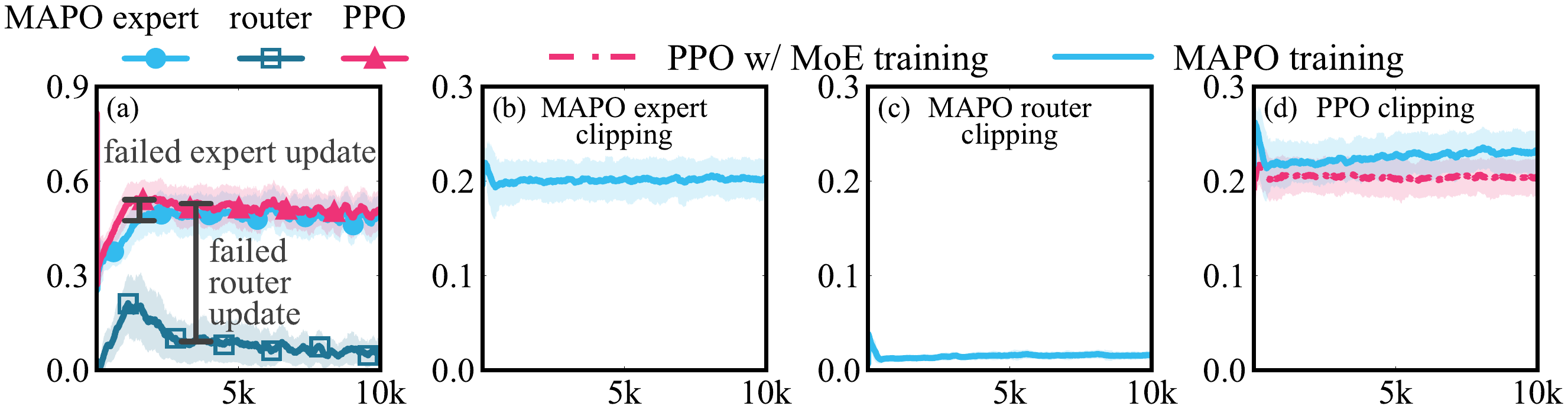}
    \caption{
    (a) Fractions of updates that would be clipped under the MAPO and PPO criteria during training with vanilla policy gradient.
    (b, c) Clipped update fractions under the MAPO expert and router criteria, respectively, during \textbf{MAPO} training.
    (d) Clipping fractions under the PPO criterion during \textbf{MAPO} training (hypothetical clipping) and \textbf{PPO with MoE} training (actual clipping).
    }
    \label{fig:clip_freq_analysis}
\end{figure}

\paragraph{Effect of Divergence Misestimation:\@}
\begin{wrapfigure}{r}{0.45\textwidth}
    \vspace{-\intextsep}
    \centering
    \captionsetup{font=small}
    \includegraphics[width=\linewidth]{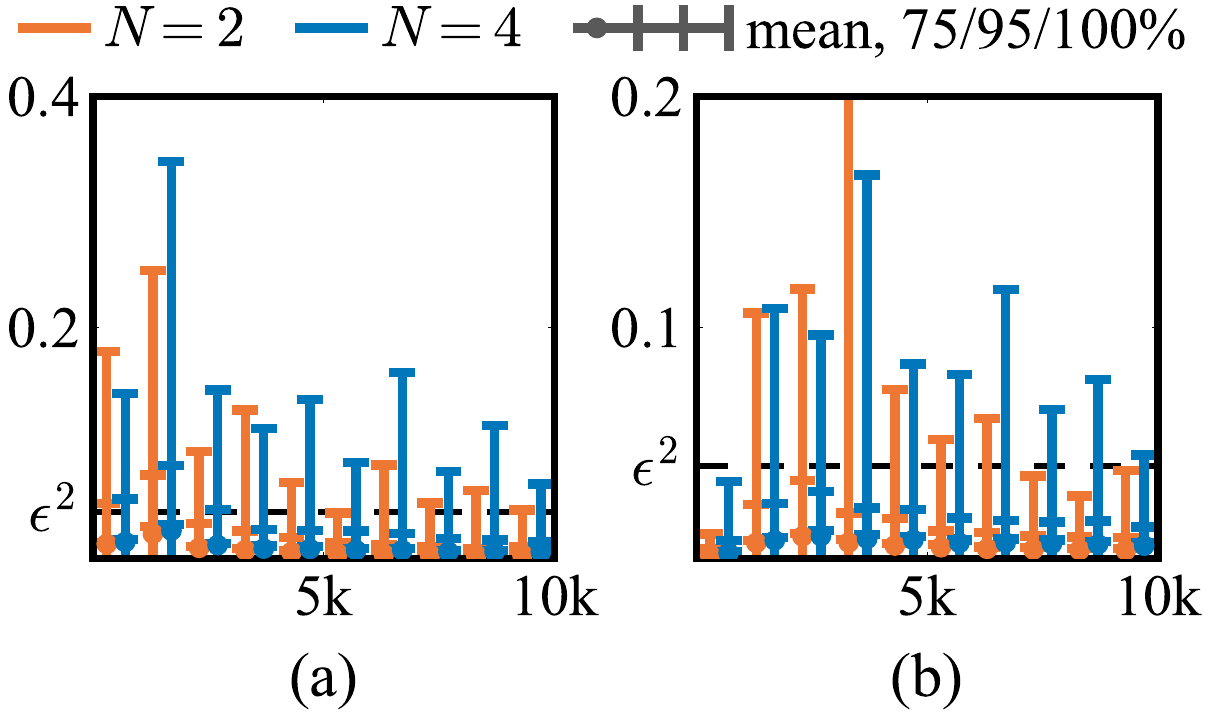}
    \vspace{-5mm}
    \caption{Mean and percentiles of divergence misestimation magnitudes. (a) Overestimation. (b) Underestimation.}
    \label{fig:misest_dist}
    \vspace{-\intextsep}
\end{wrapfigure}
We measure divergence misestimation during training with vanilla policy gradient because clipping can obscure the magnitude of misestimation.
Figure~\ref{fig:misest_dist} shows the mean and percentiles of overestimation and underestimation magnitudes.
We also evaluate the fractions of updates that would be clipped if the PPO or MAPO criteria were applied during training with vanilla policy gradient (Figure~\ref{fig:clip_freq_analysis}~(a)).
The MAPO router criterion would clip only about 5\% of updates.
In contrast, if the PPO criterion were applied to the same updates, approximately 50\% of updates would be clipped.
The higher clipping fraction for router updates is consistent with the weaker expert specialization of \textbf{PPO with MoE} (Figure~\ref{fig:total_exp}~(b, c)).

A similar pattern appears during \textbf{MAPO} training (Figure~\ref{fig:clip_freq_analysis}~(c, d)).
The MAPO router criterion clips only about 2\% of updates, compared with approximately 24\% if the PPO criterion were applied.
Notably, the PPO criterion yields similar clipping fractions during \textbf{MAPO} training and during training of \textbf{PPO with MoE} (Figure~\ref{fig:clip_freq_analysis}~(d)).
The results suggest that excessive clipping of router updates is prevented during \textbf{MAPO} training.

\paragraph{Seed Variance:\@}
The results in Table~\ref{tab:tracking_metric} and Figure~\ref{fig:total_exp} are based on a single training seed.
We further evaluate \textbf{MAPO} and \textbf{PPO with MoE} across three training seeds.

\begin{wrapfigure}{r}{0.32\textwidth}
    \centering
    \captionsetup{font=small,justification=raggedright,singlelinecheck=false}
    \includegraphics[width=\linewidth]{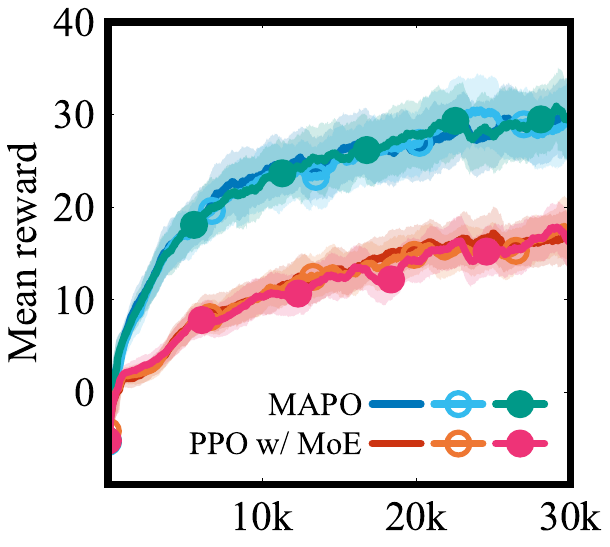}
    \vspace{-6mm}
    \caption{Learning curves of \textbf{MAPO} and \textbf{PPO with MoE} across three training seeds.}
    \label{fig:seed_sweep}
    \vspace{-\intextsep}
    \vspace{-3mm}
\end{wrapfigure}
The learning curves are shown in Figure~\ref{fig:seed_sweep}.
Across the three seeds, \textbf{MAPO} achieves a success rate of $(87.03 \pm 0.64)\%$ and a global body position error $E_{\text{g-pos}}$ of $147.9 \pm 1.5~\mathrm{mm}$.
The corresponding results for \textbf{PPO with MoE} are $(82.89 \pm 0.43)\%$ and $154.1 \pm 1.7~\mathrm{mm}$.
Success rates and tracking errors are reported as mean $\pm$ standard deviation across the three training seeds.

Additionally, we examine whether the difference in tracking success between \textbf{MAPO} and \textbf{PPO with MoE} is consistent across training seeds.
For each seed, we count motions on which \textbf{MAPO} succeeds and \textbf{PPO with MoE} fails in all five evaluation rollouts.
The counts are 182, 155, and 191 for the three training seeds, respectively.
Conversely, \textbf{PPO with MoE} consistently succeeds while \textbf{MAPO} consistently fails on only 1, 1, and 0 motions, respectively.

\end{document}